\documentclass[runningheads]{llncs}
\usepackage{cite}
\usepackage{amsmath,amssymb,amsfonts}
\usepackage{algorithmic}
\usepackage{graphicx}
\usepackage{subcaption}
\usepackage{textcomp}
\usepackage{xcolor}
\usepackage{float}
\usepackage{booktabs}
\usepackage{rotating}
\usepackage{makecell}
\usepackage{tabularx}
\usepackage{adjustbox}
\usepackage{bbm}
\usepackage{xcolor}
\usepackage{marvosym}
\definecolor{myGreen}{RGB}{0, 176, 80}
\definecolor{myBlue}{RGB}{145, 172, 224}
\definecolor{myRed}{RGB}{229, 76, 94}

\begin{document}
\title{AspectCLIP: Optimizing CLIP Representation Space via Aspect-Guided Consistency Regularization}
\titlerunning{AspectCLIP}
%
\author{Yiyang Yao \and
        Shanglin Liu \and
        Jianming Lv\textsuperscript{(\Letter)} \and
        Chengjun Wang \and
        Jinyi Li \and
        Yuchan Jie \and
        Zhihua Jin
        }
\authorrunning{Yiyang Yao et al.}


\institute{
    School of Computer Science and Technology,\\
    South China University of Technology,\\
    Guangzhou, China\\
    \email{ \{yiyang.yao.scut, slliuxjy\}@gmail.com},
    \email{jmlv@scut.edu.cn},
    \email{\{cswangchengjun, csjinyili\}@mail.scut.edu.cn},
    \email{\{jyc981214, jinzhihua0410\}@163.com}
}
%
%
%
\maketitle              

\vspace{-15pt}
\begin{abstract}
Contrastive Language-Image Pretraining learns a shared representation space through large-scale contrastive learning. However, existing methods that enforce global consistency regularization overlook a key challenge: the inherent information asymmetry between images and text: captions typically describe only one specific aspect of an image, thus images with similar visual content can be paired with completely divergent textual content and semantic information. Consequently, global regularizers inadvertently impose constraints between visually similar images whose captions describe divergent aspects, introducing semantic distortion into the representation space. We propose AspectCLIP, a framework that reformulates consistency regularization to respect this one-to-many structure. AspectCLIP first partitions training samples into attribute clusters based on textual similarity to identify aspect-coherent groups, then applies full cyclic consistency within each cluster while restricting cross-cluster regularization to prototype-level comparisons. This aspect-guided regularization enforces strict geometric alignment only when images and texts describe a consistent facet, while allowing flexibility across divergent aspects. Extensive experiments on downstream tasks demonstrate that AspectCLIP consistently outperforms traditional methods and achieves a more structured representation space.

\end{abstract}
\section{Introduction}
\label{sec:intro}

Contrastive Language-Image Pretraining (CLIP)~\cite{radford2021learning} has demonstrated remarkable success in aligning different modalities by scaling up vision-language contrastive pretraining on vast amounts of paired image-text data sourced from the web. CLIP uses two distinct encoders to map image and text data into a shared latent space, fostering a wide range of downstream applications, including zero-shot classification, zero-shot retrieval, and linear probing.

Although CLIP demonstrates outstanding performance across various tasks~\cite{sun2024alpha,wang2025iaa,yao2024minicpm}, its learning objective focuses on maximizing the similarity between matched image-text pairs relative to all mismatched image-text pairs, with no constraints imposed on the overall geometry of all samples. Consequently, it suffers from the issue of inconsistent representation space: when the same test sample is correctly classified in one modal space, it is incorrectly classified in the other modal~\cite{goel2022cyclip}.

\begin{figure}[t]
    \centering    \includegraphics[width=\linewidth]{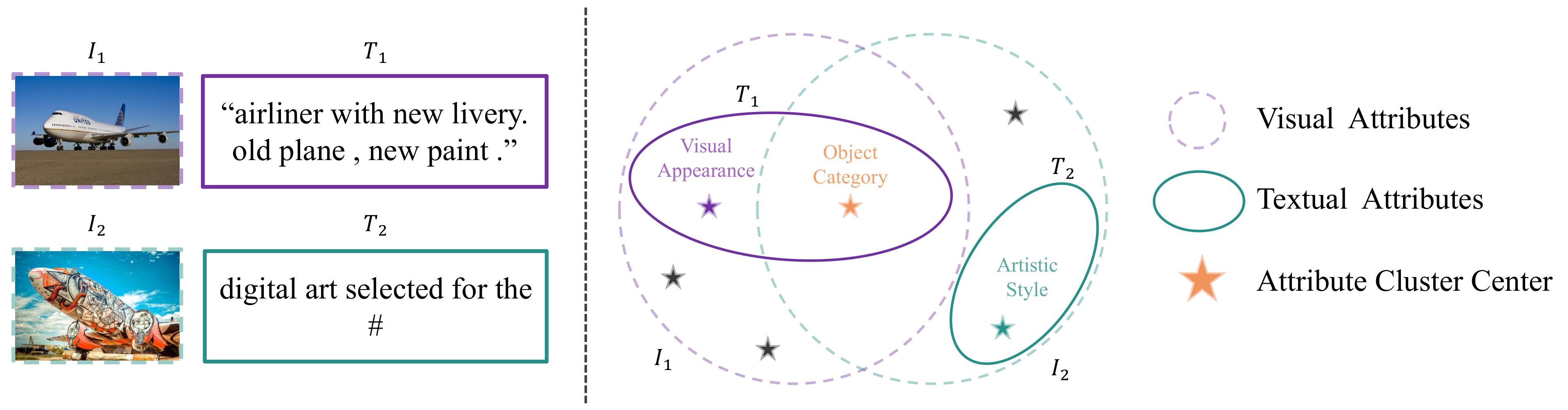}
    \caption{Illustration of the inherent information asymmetry in CLIP pretraining data. Two airplane images $(I_1, I_2)$ share the same object category but differ in visual appearance and artistic style. Their corresponding captions $(T_1, T_2)$ describe disjoint semantic aspects: $T_1$ focuses on the object's appearance and category, while $T_2$ emphasizes artistic style, demonstrating that a single image-text pair only captures a partial aspect of the visual information.}
    \label{fig:asymmetric}
    \vspace{-15pt}
\end{figure}

To address the issue of inconsistent representation space, existing research primarily falls into two directions: \textbf{Data Augmentation} and \textbf{Loss Function Optimization}. \textbf{Data Augmentation} enhances the diversity of training signals by constructing supplementary data or embedding samples~\cite{oh2023geodesic,udandarao2023sus}. However, generating suitable data requires additional generation models, which significantly elevate computational costs.
\textbf{Loss Function Optimization} enhances the geometric structure of the embedding space by adding explicit regularization terms to the loss function, without relying on external resources ~\cite{goel2022cyclip,jiang2023understanding,eslami2024mitigate}.

However, existing regularization methods largely overlook the  inherent information asymmetry of image-text pairs in CLIP pretraining data: text captions usually characterize only a single aspect of an image, rather than all its visual information. Consequently, even visually similar images may be paired with texts that carry entirely irrelevant semantic content due to their focus on distinct aspects. Specifically, a single image contains a rich spectrum of visual attributes - objects, colors, backgrounds, and more - whereas the paired short text typically describes only a small fraction of them~\cite{schrodi2024two,liu2025modfinity}. In most training samples, the caption $T$ captures only an aspect-specific subset of what the image $I$ contains, and these described aspects often differ substantially in their semantic focus~\cite{xie2025smartclip,liu2025aligning}. For example, as illustrated in Fig.\ref{fig:asymmetric}, $I_1$ and $I_2$ are two visually similar images of airplanes, which contain information about visual appearance, background, and artistic style. Their corresponding caption $T_1$ includes information about visual appearance and object category (e.g., ``airplane"), while $T_2$ focuses only on specific semantic attribute (e.g., ``artistic style"). This demonstrates the inherent one-to-many nature of visual-textual alignment in web-crawled corpora: multiple valid captions can each describe a different partial attribute of the same image, with no single caption comprehensively covering all its information.

Specifically, CyCLIP~\cite{goel2022cyclip}, a representative traditional regularization method, enhances representation stability by enforcing that in-modal distances and cross-modal distances are symmetric across the batch. However, by sampling pairs uniformly from the entire training set, these regularizers frequently pair samples that are visually similar but textually divergent in their described aspect. As illustrated in Fig.\ref{fig:cyclic}, two images might share visual similarity, but their captions could refer to entirely disjoint concepts. Forcing the distance between these images to mirror the distance between their mismatched-aspect texts introduces semantic noise into the representation learning process. 

To address this problem, we propose \textbf{AspectCLIP}, which optimizes CLIP’s representation space by accounting for the information asymmetry between images and text. Instead of applying unconstrained global regularization, AspectCLIP introduces \textbf{Aspect-Guided Consistency Regularization} to distinguish between aspect-coherent pairs and aspect-divergent pairs. Our core insight is that textual semantics define the aspect of the image being described. We first introduce an aspect-aware semantic clustering method that partitions the pretraining dataset into disjoint attribute clusters using a pretrained language model (SimCSE~\cite{gao2021simcse}), based solely on the semantic similarity of text captions. This clustering provides a structured basis for regularization: intra-cluster samples share a common descriptive aspect (e.g., captions describing ``visual appearance"), whereas inter-cluster samples diverge in their described aspects (e.g., ``visual appearance" vs. ``artistic style").

With this structure, we reformulate the regularization process by distinguishing between intra-cluster and inter-cluster consistency. Within an attribute cluster where images and texts are more likely to describe a consistent visual aspect, we enforce full cyclic consistency between individual samples to strengthen geometric alignment of that specific aspect. Across attribute clusters, where direct sample-to-sample regularization is risky due to aspect divergence, we instead enforce consistency only between individual samples and prototypes of other clusters. This design facilitates high-level, structured information exchange across aspects while naturally smoothing out the aspect-specific details that cause information conflicts. The strategy aligns geometric constraints with the natural one-to-many structure of the data: we demand strict consistency when the described aspect matches, but only loose, prototype-level consistency when aspects diverge. Extensive experiments on several downstream tasks show that AspectCLIP consistently outperforms traditional methods, while higher consistency scores confirm that our method learns a more structured representation space.

The main contributions of this paper are summarized as follows:

\begin{enumerate}
    \item We identify the inherent information asymmetry in vision-language pretraining as a key source of geometric inconsistency, and reframe global regularization as an issue of aspect divergence — rather than overlooking the one-to-many nature of the data.
    \item We propose \textbf{AspectCLIP}, an aspect-guided consistency regularization framework. By clustering text captions into attribute clusters, we decouple intra-cluster (aspect-coherent) and inter-cluster (aspect-divergent) regularization, ensuring that constraints are applied only where semantically appropriate.
    \item We validate our method through extensive experiments, achieving consistent improvements over traditional methods on both downstream task performance and representation space consistency.
\end{enumerate}

\section{Method}
\subsection{Preliminaries}
\label{sec:prelimi}
CLIP employs a dual-encoder architecture to pretrain on millions of image-text pairs crawled from the web. Consider a batch of $N$ image-text pairs $\{I_i, T_i\}_{i=1}^N$, the $i_{th}$ raw image $I_i$ and its corresponding raw text $T_i$ are input into an image encoder and a text encoder, respectively, to get the L2-normalized embeddings $\boldsymbol{v}_i, \boldsymbol{t}_i\in\mathbb{R}^d$ where $d$ is the embedding dimension. CLIP pulls the paired embeddings closer and pushes the unpaired apart through InfoNCE~\cite{oord2018representation} Loss, the image-to-text loss can be obtained as:
\begin{equation}
    \mathcal{L}_\mathrm{I2T} = -\frac{1}{N} \sum_{i=1}^N \log \frac{\exp\left[ \left( \boldsymbol{v}_i \cdot \boldsymbol{t}_i \right) / \tau \right]}{\sum_{j=1}^N \exp\left[ \left( \boldsymbol{v}_i \cdot \boldsymbol{t}_j \right) / \tau \right]},
\end{equation}
$\tau$ is a learnable temperature parameter. Text-to-image loss is defined similarly:
\begin{equation}
    \mathcal{L}_\mathrm{T2I} = -\frac{1}{N} \sum_{j=1}^N \log \frac{\exp\left[ \left( \boldsymbol{v}_j \cdot \boldsymbol{t}_j \right) / \tau \right]}{\sum_{i=1}^N \exp\left[ \left( \boldsymbol{v}_i \cdot \boldsymbol{t}_j \right) / \tau \right]}.
\end{equation}
The final CLIP loss is then:
\begin{equation}
    \mathcal{L}_\mathrm{CLIP} = \frac{1}{2} \left( \mathcal{L}_\mathrm{I2T} + \mathcal{L}_\mathrm{T2I}\right).
\end{equation}

To quantify the cross-modal alignment, CyCLIP~\cite{goel2022cyclip} formally defines the \textbf{Consistency Score} that measures the agreement between predicted labels derived from the text and image spaces, respectively. Specifically, in the text embedding space, the label of a test image ($P_T$) is predicted under zero-shot setting by selecting the text label with the highest cosine similarity. In the image embedding space, given a labeled training set, the predicted label ($P_I^k$) is obtained via a majority vote over the true labels of the test image’s $k$-nearest training samples. Formally, the consistency score is defined as:
\begin{equation}
    \text{Consistency\ Score}_k = \frac{1}{N} \sum_{i=1}^N \mathbbm{1}[ P_I^k(I_i) = P_T(I_i) ],
\end{equation}
$N$ is the number of test images and $\mathbbm{1}[\cdot]$ is indicator function.
\vspace{-10pt}

\begin{figure}[t]
  \centering
  \includegraphics[width=0.8\linewidth]{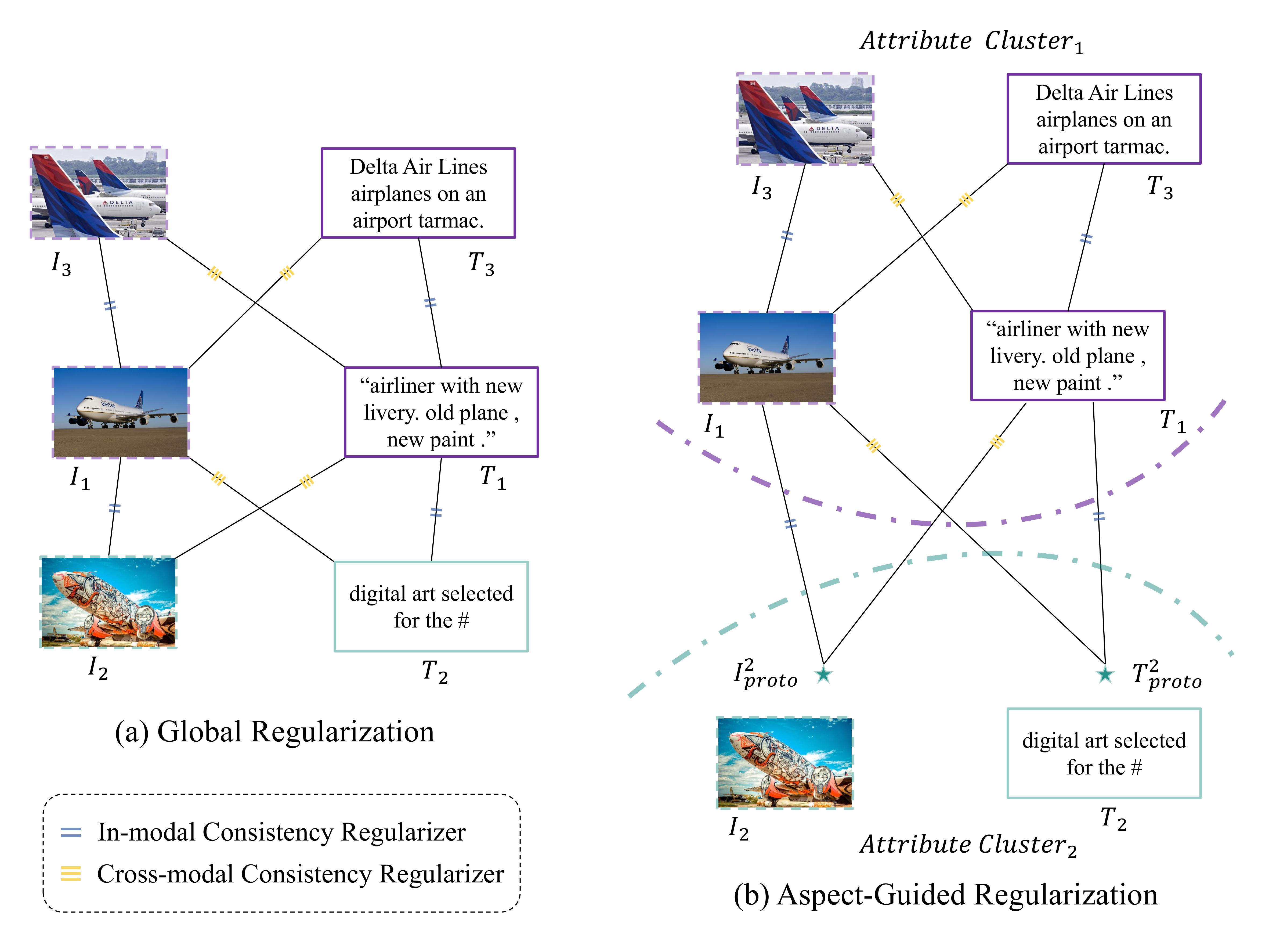}
  \caption{An illustration of global regularization without constraints (CyCLIP) and our aspect-guided regularization. Edges indicate the distance between embeddings \textit{i.e.,} $d(e_1,e_2)$. CyCLIP ensures cyclic consistency in the representation space by aligning in-modal distances $(d(I_1,I_2) \sim d(T_1,T_2))$ and cross-modal distances $(d(I_1,T_2) \sim d(T_1,I_2))$. Our aspect-guided regularization further optimizes this by taking information asymmetry between modalities into consideration.}
  \label{fig:cyclic}
  \vspace{-10pt}
\end{figure}

\subsection{Aspect-aware Semantic Clustering on Pretraining Data}
\label{sec:cluster}
As illustrated in Fig.~\ref{fig:cyclic}, traditional regularization methods such as CyCLIP enforce regularizers on any two pairs from the global pretraining set. The two cyclic consistency regularizers added by CyCLIP can be formulated as:
\begin{enumerate}
    \item The cross-modal consistency regularizer aligns the similarity scores between the embeddings of every mismatched image-text pair within a batch:
    \begin{equation}
        \mathcal{L}_\mathrm{C-Cyclic} = \frac{1}{N} \sum_{i=1}^N \sum_{j=1}^N \left( \boldsymbol{v}_i \cdot \boldsymbol{t}_j - \boldsymbol{v}_j \cdot \boldsymbol{t}_i \right)^2.
    \end{equation}

    \item The in-modal consistency regularizer minimizes differences in the similarity scores between the embeddings of all possible image-image pairings and their corresponding text-text pairings within a single batch:
    \begin{equation}
        \mathcal{L}_\mathrm{I-Cyclic} = \frac{1}{N} \sum_{i=1}^N \sum_{j=1}^N \left( \boldsymbol{v}_i \cdot \boldsymbol{v}_j - \boldsymbol{t}_j \cdot \boldsymbol{t}_i \right)^2.
    \end{equation}
\end{enumerate}

However, the global regularization in CyCLIP overlooks the inherent information asymmetry between images and text. As shown in Fig.~\ref{fig:cyclic}, $(I_1,T_1)$ is an image-text pair about an airliner with new livery; $(I_2,T_2)$ is another pair where $I_2$ depicts an old airplane parked outdoors with graffiti-style livery, and $T_2$ is ``\textit{digital art selected for the \#}", which describes only a limited subset of the visual content in $I_2$. Here, $I_1$ and $I_2$ are two visually similar images, but their captions $T_1$ and $T_2$ diverge significantly in their described aspects. Meanwhile, $(I_2,T_1)$ actually shares a more coherent descriptive aspect than $(I_1,T_2)$, because $T_1$ happens to capture a visual facet also present in $I_2$. CyCLIP, however, treats all cross-pair relationships equally and enforces regularizers to make in-modal distances $(d(I_1,I_2) \sim d(T_1,T_2))$ and cross-modal distances $(d(I_1,T_2) \sim d(T_1,I_2))$ symmetric. This forces the model to reconcile a textual divergence that stems from a legitimate aspectual difference rather than a visual error, undermining the effectiveness of the regularization.

\begin{figure*}[t]
  \centering
  \includegraphics[width=\linewidth]{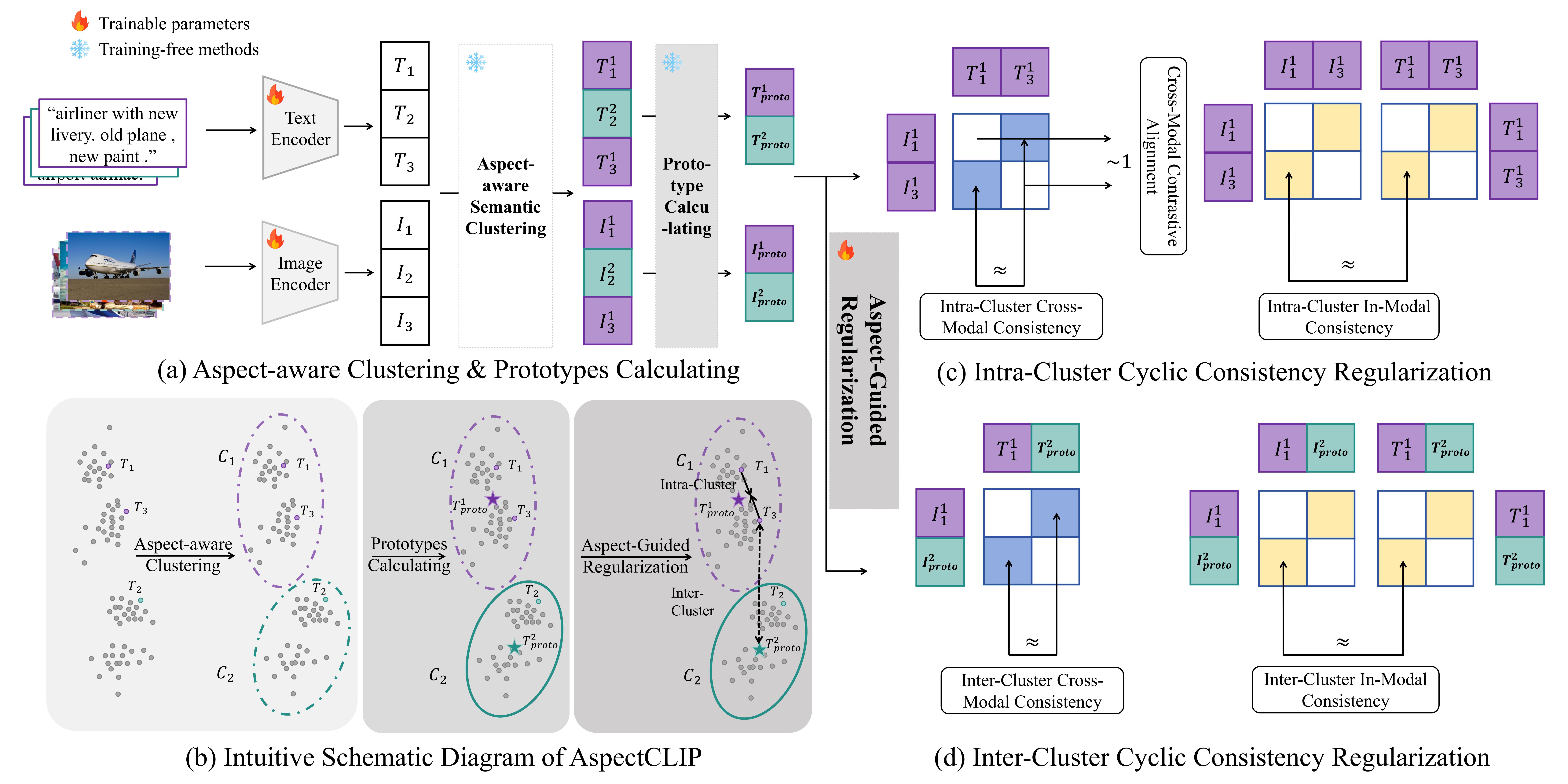}
  \caption{The overall framework of AspectCLIP. We first introduce aspect-aware semantic clustering to partition the pretraining dataset into disjoint attribute clusters, then enforce aspect-guided consistency regularization based on these clusters to optimize the CLIP representation space by aligning geometric constraints with the natural one-to-many structure of visual-textual descriptions.}
  \label{fig:framework}
  \vspace{-15pt}
\end{figure*}

To solve the problems mentioned above, we introduce an aspect-aware semantic clustering method that adaptively partitions pretraining data into distinct attribute semantic clusters, where intra-cluster samples share a common descriptive aspect and inter-cluster samples describe divergent aspects. Inspired by MoDE~\cite{ma2024mode}, we leverage the semantic similarity inherent in text captions: let the full pretraining set be $\{I_i,T_i\}_{i=1}^M$ where $M$ denotes the total number of pretraining samples, we first extract text embeddings via a pretrained language model SimCSE~\cite{gao2021simcse}, then apply a \textit{K-Means clustering} on text embeddings to partition the set into $K$ disjoint attribute clusters ${C_1, C_2,..., C_K}$. Each cluster $C_k$ contains a subset of samples, denoted as:
\begin{equation}
C_k = \{ (I_i^k, T_i^k) \mid i=1,2,...,N_k \},
\end{equation}
where $N_k$ is the number of samples in cluster $C_k$, and $\sum{k=1}^K N_k=M$. The superscript $k$ indicates the cluster affiliation of the sample $(I_i, T_i)$.

After clustering, $(I_1,T_1)$ and $(I_2,T_2)$ are unlikely to appear in the same attribute cluster as $T_1$ and $T_2$ describe semantically divergent aspects. The clustering provides a structured attribute-level basis for targeted regularization; the disjoint attribute clusters ${C_1, C_2,..., C_K}$ act as interpretable units that distinguish aspect-coherent pairs from aspect-divergent pairs in subsequent regularization.

Our decision to cluster solely on text captions is motivated by the fundamental information asymmetry in vision-language data. In a typical image-text pair, the text captures only a small fraction of the information in the image — it describes a specific aspect, such as category, color, or style, while the image encodes a far richer spectrum of visual information~\cite{xie2025smartclip}. Consequently, similar captions almost certainly describe the same aspect of their corresponding images, so the images likely share visual content along that aspect. Conversely, visually similar images may have captions describing entirely different attributes, and thus would not belong to the same attribute cluster in terms of aspects~\cite{ma2024mode}. Thus, text-based clustering more reliably identifies aspect-coherent groups, whereas image-based clustering risks mixing samples with divergent aspects.

Note that although our aspect-aware clustering is implemented similarly to MoDE~\cite{ma2024mode}, they differ fundamentally in their core objective and downstream application. In MoDE, the clustering step is designed to partition the pretraining data into distinct groups so that each cluster can be assigned to a dedicated ``data expert" (a specialized sub-model) for fine-grained pretraining. In contrast, our aspect-aware clustering serves as a structural basis for distinguishing attribute aspects, enabling the regularization to target aspect-coherent and aspect-divergent pairs differently.
\vspace{-10pt}

\subsection{Intra-cluster Consistency Regularization}
\label{sec:intra}
The aspect-aware clustering partitions pretraining data into distinct attribute semantic clusters, where intra-cluster samples are more likely to describe a consistent visual aspect. Our aspect-guided regularization thus enforces full cyclic consistency between individual samples within each cluster to strengthen the geometric alignment of that shared aspect.

Mathematically, we first define the intra-cluster sample pair set within the current batch as:
\begin{equation}
    \mathcal{S}=\{ ((I_i^k, T_i^k),(I_j^k, T_j^k)) \mid (I_i^k, T_i^k) \in C_k, (I_j^k, T_j^k) \in C_k \},
\end{equation}
where $(I_i^k, T_i^k)$ and $(I_j^k, T_j^k)$ are two image-text pairs from the same cluster $C_k$. The size of $\mathcal{S}$ is the total number of intra-cluster pairs across all clusters in the current batch:
 \begin{equation}
     |\mathcal{S}|=\sum_{k=1}^K(N_k^b)^2,
 \end{equation}
where $N_k^b$ denotes the number of samples from cluster $k$ in batch $b$. If we denote the total number of batches as $B$ and batch size as $N$, then $\sum_{k=1}^K N_k^b = N$ and $\sum_{b=1}^B N_k^b = N_k$. The intra-cluster cyclic consistency regularizers are given as:

\begin{equation}
\mathcal{L}_\mathrm{C-Cyclic}^\text{intra} = \frac{1}{|\mathcal{S}|} \sum_{((I_i^k, T_i^k),(I_j^k, T_j^k)) \in \mathcal{S}} \left( \boldsymbol{v}_i^k \cdot \boldsymbol{t}_j^k - \boldsymbol{v}_j^k \cdot \boldsymbol{t}_i^k \right)^2.
\end{equation}

\begin{equation}
\mathcal{L}_\mathrm{I-Cyclic}^\text{intra} = \frac{1}{|\mathcal{S}|} \sum_{((I_i^k, T_i^k),(I_j^k, T_j^k)) \in \mathcal{S}} \left( \boldsymbol{v}_i^k \cdot \boldsymbol{v}_j^k - \boldsymbol{t}_j^k \cdot \boldsymbol{t}_i^k \right)^2,
\end{equation}
The total intra-cluster consistency loss is the aggregation:
\begin{equation}
\mathcal{L}_{\text{intra}}=\mathcal{L}_\mathrm{I-Cyclic}^\text{intra} + \mathcal{L}_\mathrm{C-Cyclic}^\text{intra}
\end{equation}

\subsection{Inter-cluster Consistency Regularization}
\label{sec:inter}
With aspect-aware clustering, we consider inter-cluster samples as describing divergent visual aspects. Our aspect-guided regularization enforces consistency only between individual samples and the prototypes of other clusters. This encourages high-level, structured information exchange across aspects while naturally smoothing out aspect-specific details that cause the information asymmetry.

Mathematically, we first compute the image and text prototypes of each cluster $k$ as the mean embeddings of all samples within the cluster in the current batch $b$:
\begin{equation}
\boldsymbol{v}_\text{proto}^k = \frac{1}{N_k^b} \sum_{i=1}^{N_k^b} \boldsymbol{v}_i^k, \quad
\boldsymbol{t}_\text{proto}^k = \frac{1}{N_k^b} \sum_{i=1}^{N_k^b} \boldsymbol{t}_i^k,
\end{equation}
where $\boldsymbol{v}_i^k$ and $\boldsymbol{t}_i^k$ denote the $i$-th image and text embeddings in cluster $k$, respectively. These prototypes capture core semantic features of each attribute aspect and naturally smooth out interference from individual aspect-specific variations.

Next, the inter-cluster cyclic consistency regularizers are added between each sample in cluster $k$ and the prototypes of all other clusters $m\neq k$:
\begin{equation}
\mathcal{L}_{\mathrm{C-Cyclic}}^{\text{inter},k,i} = \frac{1}{K-1} \sum_{\substack{m=1 \\ m \neq k}}^K \left( \boldsymbol{v}_i^k \cdot \boldsymbol{t}_\text{proto}^m - \boldsymbol{v}_\text{proto}^m \cdot \boldsymbol{t}_i^k \right)^2,
\end{equation}

\begin{equation}
\mathcal{L}_{\mathrm{I-Cyclic}}^{\text{inter},k,i} = \frac{1}{K-1} \sum_{\substack{m=1 \\ m \neq k}}^K \left( \boldsymbol{v}_i^k \cdot \boldsymbol{v}_\text{proto}^m - \boldsymbol{t}_\text{proto}^m \cdot \boldsymbol{t}_i^k  \right)^2.
\end{equation}
Aggregating over all samples and all clusters in the batch, the total inter-cluster consistency regularization loss is:
\begin{equation}
\mathcal{L}_{\text{inter}} = \frac{1}{N} \sum_{k=1}^K \sum_{i=1}^{N_k^b} \left( \mathcal{L}_{\mathrm{C-Cyclic}}^{\text{inter},k,i} + \mathcal{L}_{\mathrm{I-Cyclic}}^{\text{inter},k,i} \right).
\end{equation}

\subsection{Total Loss}
\label{sec:total}
AspectCLIP enforces aspect-guided consistency regularization via aspect-aware clustering to optimize the CLIP representation space under the natural one-to-many structure of visual-textual descriptions. The final training objective integrates the original CLIP contrastive loss, the intra-cluster consistency regularization loss, and the inter-cluster consistency regularization loss:
\begin{equation}
\mathcal{L}_{\text{total}} = \mathcal{L}_{\text{CLIP}} + \lambda_1 \cdot \mathcal{L}_{\text{intra}} + \lambda_2 \cdot \mathcal{L}_{\text{inter}},
\end{equation}
where $\lambda_1$ and $\lambda_2$ are hyperparameters that control the contribution of intra-cluster and inter-cluster regularization.

Note that we still compute the CLIP contrastive loss on the global collection of pretraining samples since the diversity of negatives is the key to the success of contrastive learning.

\vspace{-10pt}

\section{Experiments and Analysis}

\subsection{Pretraining Dataset and Setup}
We use the CC3M~\cite{sharma2018conceptual} dataset for pretraining all the models, which has been used as a benchmark in many subsequent works~\cite{goel2022cyclip,gao2024softclip} of CLIP. We pretrain our model from scratch for 64 epochs,
for fair comparison, we also pretrain CLIP~\cite{radford2021learning} and CyCLIP\cite{goel2022cyclip} from scratch for 64 epochs. For zero-shot image classification and robustness to natural distribution shifts, we additionally report results from LMS~\cite{jiang2023understanding} for comparison, as their work adopts the same pretraining setup as ours. Since their paper reports results for three model variants, we average them.

The experiments are conducted on 2 A800 GPUs with a batch size of 128; all models use ResNet-50 as image encoder and a transformer architecture as text encoder. We use $\lambda_1=0.25$, $\lambda_2=0.25$ and $K=4$ throughout our experiments. For testing the efficacy of both intra-cluster and inter-cluster consistency regularization, we report two variants of our model, where \textbf{AspectCLIP} refers to the model with both regularization and \textbf{AspectCLIP(Intra)} denotes the model with intra-cluster regularization only.

We note that SimCSE~\cite{gao2021simcse} is used as an off-the-shelf model without any additional training. Aspect-aware semantic clustering, performed once as a static pre‑processing step before pretraining, has a time complexity of $O(I \times n \times k \times m)$ and a space complexity of $O(n \times m)$, where $n$ is the number of samples, $m$ the embedding dimension, $k$ the number of clusters, and $I$ the number of iterations. Both costs scale linearly with the dataset size, making the approach easily extensible to larger corpora such as LAION‑400M or CC12M.
\vspace{-10pt}

\begin{table}[t] 
    \centering
    \caption{Zero-shot classification accuracy (\%)}
    \label{tab:zero_shot_cls_acc}
    \begin{tabular}{lccccccccc} 
        \toprule 
        & \multicolumn{3}{c}{CIFAR-10} & \multicolumn{3}{c}{CIFAR-100} & \multicolumn{3}{c}{ImageNet1K}\\
        \cmidrule(lr){2-4} \cmidrule(lr){5-7} \cmidrule(lr){8-10} 
         & Top1 & Top3 & Top5 & Top1 & Top3 & Top5 & Top1 & Top3 & Top5\\
        \midrule 
        CLIP & 46.50 & 77.98 & 89.06 & 18.67 & 34.58 & 43.49 & 19.92 & 32.71 & 39.14\\
        LMS & 48.71 & 78.15 & 89.81 & 18.76 & 34.93 & 43.77 & 20.30 & 33.44 & 39.60\\
        CyCLIP & 51.38 & 78.67 & 89.44 & 23.39 & 41.27 & 49.93 & 21.50 & 35.03 & \underline{41.42}\\
        \midrule 
        \textbf{AspectCLIP(Intra)} & \underline{51.72} & \underline{81.14} & \underline{91.79} & \underline{24.84} & \textbf{43.94} & \textbf{53.48} & \underline{21.71} & \underline{35.10} & 41.24\\
        \textbf{AspectCLIP} & \textbf{52.06} & \textbf{81.42} & \textbf{92.26} & \textbf{25.22} & \underline{42.91} & \underline{52.88} & \textbf{21.87} & \textbf{35.33} & \textbf{41.69}\\
        \bottomrule 
    \end{tabular}
    \vspace{-10pt}
\end{table}

\begin{table}[t] 
    \centering
    \caption{Zero-shot classification on natural distribution shifts (\%)}
    \label{tab:zero_shot_nds_acc}
    \begin{tabular}{lccccccccc} 
        \toprule 
        & \multicolumn{3}{c}{ImageNetV2} & \multicolumn{3}{c}{ImageNetSketch} & \multicolumn{3}{c}{ImageNet-R}\\
        \cmidrule(lr){2-4} \cmidrule(lr){5-7} \cmidrule(lr){8-10}
         & Top1 & Top3 & Top5 & Top1 & Top3 & Top5 & Top1 & Top3 & Top5\\
        \midrule 
        CLIP & 18.14 & 30.96 & 35.52 & 9.79 & 17.50 & 23.23 & 22.35 & 35.68 & 43.09 \\
        LMS & 17.06 & 29.40 & 35.95 & 10.15 & 19.01 & 24.05 & 22.10 & 35.33 & 42.16 \\
        CyCLIP & 21.05 & 35.31 & 40.72 & \underline{11.66} & \underline{21.09} & \underline{26.46} & 24.88 & 40.13 & 47.67 \\
        \midrule 
        \textbf{AspectCLIP(intra)} & \underline{24.13} & \underline{38.89} & \underline{45.05} & 11.20 & 20.52 & 25.89 & \underline{25.86} & \underline{41.34} & \underline{48.90}\\
        \textbf{AspectCLIP} & \textbf{24.18} & \textbf{39.19} & \textbf{45.38} & \textbf{12.05} & \textbf{21.62} & \textbf{26.96} & \textbf{26.07} & \textbf{42.11} & \textbf{49.85}\\
        \bottomrule 
    \end{tabular}
    \vspace{-10pt}
\end{table}

\subsection{Zero-Shot Image Classification}
We conduct the zero-shot image classification on CIFAR-10, CIFAR-100, and ImageNet1K with the combination of text prompts used by CLIP, \textit{e.g.,} ``a photo of the \{class name\}". The experimental results are summarized in  Table~\ref{tab:zero_shot_cls_acc}. We can see from the results that AspectCLIP achieves consistent improvement over CLIP, LMS, and CyCLIP, across the three benchmarks. Specifically, AspectCLIP(Intra) achieves a 3\% and 6\% Top-3 accuracy improvement over CyCLIP on CIFAR-10 and CIFAR-100 datasets, respectively, and remains competitive with CyCLIP on ImageNet1K dataset. AspectCLIP further improves zero-shot performance on CIFAR-10 and ImageNet1K, while exhibits a slight decline on CIFAR-100. Our experiments evince that aspect-guided consistency regularization effectively enhances zero-shot classification by leveraging the intrinsic structure of descriptive aspects in pretraining data.

\vspace{-25pt}

\begin{table}[htbp]
  \centering
  \setlength{\tabcolsep}{5pt}
  \caption{Accuracy scores for image classification with linear probing (\%)}
  \label{tab:transfer_1}
  \small
  \begin{tabularx}{\linewidth}{l *{12}{c} c}  
    \toprule
    & 
    \makecell{Caltech101} & 
    \makecell{CIFAR-10} & 
    \makecell{CIFAR-100} & 
    \makecell{DTD} &   
    \makecell{GTSRB}\\
    \midrule
    CLIP & 
    80.97 & \textbf{77.76} & 52.91 & 60.13 & 64.38\\
    CyCLIP & 
    81.89 & 76.44 & 53.59 & 61.97 & \underline{66.91}\\
    \midrule
    \textbf{AspectCLIP(Intra)} &
    \textbf{82.19} & \underline{76.68} & \underline{54.13} & \underline{63.62} & \textbf{68.80}\\
    \textbf{AspectCLIP} &
    \underline{82.04} & 76.61 & \textbf{54.20} & \textbf{64.10} & 65.86\\
    \bottomrule
  \end{tabularx}
  \vspace{-15pt}
\end{table}

\begin{table}[htbp]
  \centering
  \setlength{\tabcolsep}{5pt}
  \caption{Accuracy scores for image classification with linear probing (\%)}
  \label{tab:transfer_2}
  \small
  \begin{tabularx}{\linewidth}{l *{12}{c} c}  
    \toprule
    & 
    \makecell{ImageNet1K} & 
    \makecell{OxfordPets} &
    \makecell{SST2} & 
    \makecell{STL10} & 
    \makecell{SVHN}\\
    \midrule
    CLIP & 
    35.52 & \underline{57.26} & 54.07 & 89.28 & 46.25\\
    CyCLIP & 
    \textbf{36.94} & 56.99 & 54.31 & 90.15 & 46.98\\
    \midrule
    \textbf{AspectCLIP(Intra)} &
    \underline{36.89} & 56.69 & \textbf{55.68} & \underline{90.61} & \textbf{47.85}\\
    \textbf{AspectCLIP} &
    36.81 & \textbf{57.35} & \underline{55.35} & \textbf{90.74} &  \underline{47.69}\\
    \bottomrule
  \end{tabularx}
  \vspace{-15pt}
\end{table}

\vspace{-20pt}
\subsection{Robustness to Natural Distribution Shifts}
On zero-shot image classification, CLIP further demonstrates impressive robustness when the test image distribution shifts to natural variants such as sketches, cartoons, or renditions. In Table~\ref{tab:zero_shot_nds_acc}, we investigate the performance improvement brought by AspectCLIP when facing natural distribution shifts. The results on three benchmarks: ImageNetV2~\cite{recht2019imagenet}, ImageNetSketch~\cite{wang2019learning} and ImageNet-R~\cite{hendrycks2021many} show that AspectCLIP achieves obvious improvement over previous methods. The enhancement is particularly impressive in the ImageNetV2 dataset, where we achieve a 15\% Top-1 accuracy improvement and 11\% Top-3 accuracy improvement. Based on the observations, we conclude that respecting the one-to-many nature of image-text descriptions through aspect-guided regularization significantly strengthens CLIP's robustness to natural distribution shifts.
\vspace{-11pt}

\subsection{Linear Probing}
We further evaluate the learned representations via linear probing: we freeze the pretrained visual encoder (ResNet-50) and fit a linear classifier on its output embeddings using the labeled training sets of several image classification benchmarks. As shown in Table~\ref{tab:transfer_1} and Table~\ref{tab:transfer_2}, AspectCLIP outperforms the baselines on eight out of ten benchmarks, demonstrating that aspect-guided consistency regularization also strengthens transferable features under supervised evaluation.
\vspace{-20pt}

\subsection{Consistency in Representation Space}
As we introduced in Section~\ref{sec:prelimi}, the consistency score serves as a quantitative measurement of cross-modal alignment. We test the consistency scores on four benchmarks and report the results in Table~\ref{tab:cons_score_1} and Table~\ref{tab:cons_score_2}. The consistency score is calculated over 6K, 10K, 10K, and 50K testing images of Caltech101, CIFAR-10, CIFAR-100 and ImageNet1K dataset, respectively. AspectCLIP achieves obvious improvement on Caltech101 and CIFAR-100, and is slightly better on CIFAR-10 and ImageNet1K. We conclude that while AspectCLIP delivers stronger performance on several downstream tasks, it also learns a more consistent representation space, owing to its aspect-guided regularization that respects the one-to-many nature of image-text descriptions.
\vspace{-10pt}

\begin{table}[t] 
    \centering
    \setlength{\tabcolsep}{4pt} 
    \caption{Consistency score (\%) trend across standard benchmarks}
    \label{tab:cons_score_1}
    \begin{tabular}{lcccccccc} 
        \toprule
        & \multicolumn{4}{c}{Caltech101} & 
         \multicolumn{4}{c}{CIFAR-10}\\
        \cmidrule(lr){2-5} \cmidrule(lr){6-9} 
         & Top1 & Top3 & Top5 & Top10 & Top1 & Top3 & Top5 & Top10\\
        \midrule
        CLIP & 48.40 & 49.45 & 49.89 & 49.67 & 44.84 & 47.05 & 47.92 & 48.59  \\
        CyCLIP & 52.07 & 52.68 & 53.30 & 53.67 & 48.94 & \underline{50.88} & \underline{52.51} & \textbf{53.95}\\
        \midrule
        \textbf{AspectCLIP(Intra)} & \textbf{54.13} & \textbf{55.25} & \textbf{55.87} & \textbf{56.33} & \underline{49.21} & \textbf{51.39} & \textbf{52.67} & 53.19\\
        \textbf{AspectCLIP} & \underline{53.18} & \underline{53.85} & \underline{54.21} & \underline{54.85} & \textbf{49.42} & 50.59 & 52.30 & \underline{53.37}\\
        \bottomrule
    \end{tabular}
    \vspace{-10pt}
\end{table}

\begin{table}[t] 
    \centering
    \setlength{\tabcolsep}{4pt} 
    \caption{Consistency score (\%) trend across standard benchmarks}
    \label{tab:cons_score_2}
    \begin{tabular}{lcccccccc} 
        \toprule
        & \multicolumn{4}{c}{CIFAR-100} & \multicolumn{4}{c}{ImageNet1K} \\
        \cmidrule(lr){2-5} \cmidrule(lr){6-9} 
         & Top1 & Top3 & Top5 & Top10 & Top1 & Top3 & Top5 & Top10 \\
        \midrule
        CLIP & 16.08 & 16.99 & 18.21 & 19.15 & 16.30 & 17.06 & 18.27 & 19.54  \\
        CyCLIP & 20.88 & 21.43 & 22.73 & 24.56 & 19.19 & 19.82 & 21.60 & 23.64\\
        \midrule
        \textbf{AspectCLIP(Intra)} &  \textbf{22.59} & \textbf{23.40} & \underline{24.73} & \underline{26.40} & \underline{19.26} & \underline{19.89} & \underline{21.70} & \underline{23.82}\\
        \textbf{AspectCLIP} & \underline{22.36} & \underline{23.39} & \textbf{25.11} & \textbf{26.72} & \textbf{19.43} & \textbf{20.04} & \textbf{21.85} & \textbf{24.16} \\
        \bottomrule
    \end{tabular}
    \vspace{-10pt}
\end{table}

\begin{figure}[t]
  \centering
  \includegraphics[width=0.95\linewidth]{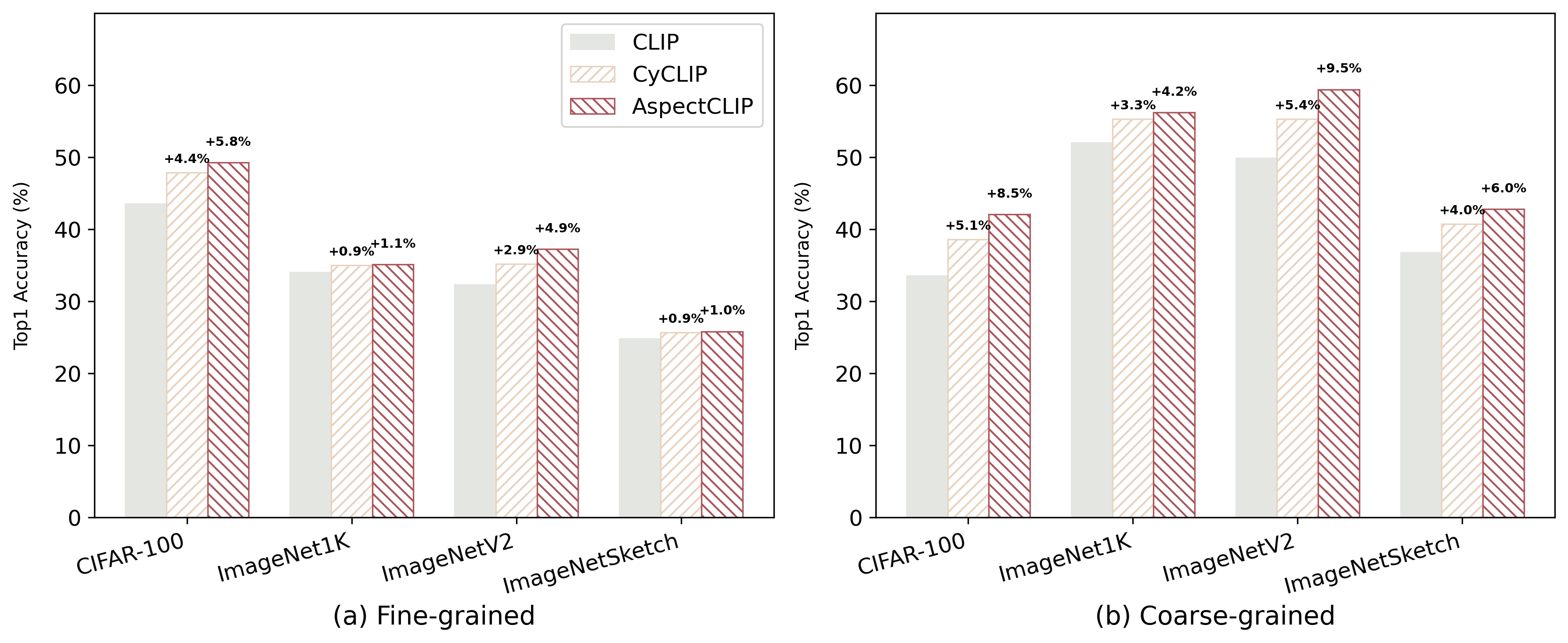}
  \caption{Fine-grained and coarse-grained zero-shot classification performance (reported gains are relative to CLIP)}
  \label{fig:fine_coarse}
  \vspace{-15pt}
\end{figure}

\subsection{Fine-grained and Coarse-grained Performance}
We follow CyCLIP to investigate how different methods perform on coarse- and fine-grained zero-shot classification. Given a test set of $N$ ``image-subclass-superclass" triplets $\{I_j, C_j, P_j\}_{j=1}^N$, a superclass $P_j$ denotes the coarse-grained, high-level category (e.g., ``big cats") and a subclass $C_j$ denotes the fine-grained, low-level category (e.g., ``lion", ``tiger", ``leopard"). We define \textbf{Fine-grained accuracy} as the proportion of cases where the top-matching subclass for an image exactly equals its true subclass $C_j$; and \textbf{Coarse-grained accuracy} as the proportion of cases where the predicted subclass belongs to the set of subclasses under the image's true superclass $P_j$, even if the exact subclass is incorrect.

As shown in Fig.~\ref{fig:fine_coarse}, AspectCLIP consistently outperforms traditional methods on both metrics, with the improvement more pronounced for coarse‑grained than fine‑grained classification across all datasets. We attribute this to our aspect‑guided regularization: intra‑cluster consistency sharpens fine‑grained distinctions within the same aspect, while inter‑cluster prototype‑based regularization learns clearer high‑level boundaries across aspects. Since coarse‑grained categories often align with attribute‑level semantic divisions (e.g., visual appearance vs. artistic style), the clustering provides an inductive bias that particularly benefits coarse‑grained discrimination. This suggests aspect‑guided consistency regularization is inherently well‑suited for learning concept‑level representations.

\begin{table}[t]  
    \footnotesize  
    \centering
    \caption{Zero-shot classification accuracy (\%)}
    \label{tab:hyper_analy_zs}
    \resizebox{\linewidth}{!}{
    \begin{tabular}{lccccccccc}
        \toprule
        & \multicolumn{3}{c}{CIFAR-10} & \multicolumn{3}{c}{CIFAR-100} & \multicolumn{3}{c}{ImageNet1K} \\
        \cmidrule(lr){2-4} \cmidrule(lr){5-7} \cmidrule(lr){8-10} 
        Method & Top1 & Top3 & Top5 & Top1 & Top3 & Top5 & Top1 & Top3 & Top5\\
        \midrule
        CyCLIP & 51.38 & 78.67 & 89.44 & 23.39 & 41.27 & 49.93 & 21.50 & 35.03 & \underline{41.42}\\
        \midrule
        \textbf{AspectCLIP($\lambda_2=0.1$)} & 51.52 & \underline{80.31} & 91.05 & \underline{24.67} & \textbf{43.02} & \underline{52.79} & \textbf{21.91} & \textbf{35.46} & \textbf{41.84}\\
        \textbf{AspectCLIP($\lambda_2=0.25$)} & \textbf{52.06} & \textbf{81.42} & \textbf{92.26} & \textbf{25.22} & \underline{42.91} & \textbf{52.88} & \underline{21.87} & \underline{35.33} & \underline{41.69}\\
        \textbf{AspectCLIP($\lambda_2=0.5$)} & \underline{51.63} & 79.78 & \underline{91.27} & 24.20 & 41.93 & 51.75 & 21.77 & 35.22 & 41.34 \\
        \bottomrule
    \end{tabular}
    }
    \vspace{-10pt}
\end{table}
\vspace{-10pt}

\begin{table}[t]  
    \footnotesize  
    \centering
    \caption{Zero-shot classification on natural distribution shifts (\%)}
    \label{tab:hyper_analy_nds}
    \resizebox{0.75\linewidth}{!}{
    \begin{tabular}{lcccccc}
        \toprule
        & \multicolumn{3}{c}{ImageNetV2} & \multicolumn{3}{c}{ImageNetSketch}\\
        \cmidrule(lr){2-4} \cmidrule(lr){5-7}  
        Method & Top1 & Top3 & Top5 & Top1 & Top3 & Top5\\
        \midrule
        CyCLIP & 21.05 & 35.31 & 40.72 & 11.66 & 21.09 & 26.46 \\
        \midrule
        \textbf{AspectCLIP($\lambda_2=0.1$)} & \underline{24.14} & \underline{38.72} & \underline{45.22} & \underline{11.67} & \underline{21.45} & \underline{26.86} \\
        \textbf{AspectCLIP($\lambda_2=0.25$)} & \textbf{24.18} & \textbf{39.19} & \textbf{45.38} & \textbf{12.05} & \textbf{21.62} & \textbf{26.96} \\
        \textbf{AspectCLIP($\lambda_2=0.5$)} & 23.18 & 37.92 & 44.41 & 10.32 & 18.97 & 23.04 \\
        \bottomrule
    \end{tabular}
    }
    \vspace{-10pt}
\end{table}

\subsection{Hyperparameters analysis}
\vspace{-5pt}
We empirically set $K=4$ following MoDE~\cite{ma2024mode} and $\lambda_1=0.25$ following CyCLIP~\cite{goel2022cyclip}. For $\lambda_2$, we conduct experiments to explore its effect on zero-shot classification and natural distribution shifts, as shown in Table~\ref{tab:hyper_analy_zs} and Table~\ref{tab:hyper_analy_nds}. We evaluate three settings with $\lambda_2$ set to $0.1$, $0.25$ and $0.5$.  
A relatively small $\lambda_2=0.1$ yields competitive results on both standard and shifted datasets. Increasing $\lambda_2$ to $0.25$, AspectCLIP achieves the best overall accuracy across nearly all metrics, demonstrating optimal balance for aspect-aware alignment. However, further enlarging $\lambda_2$ to $0.5$ causes clear performance drops, especially on distribution shifts, indicating that an overly large coefficient over-constrains the aspect learning branch and harms generalization. Thus, we choose $\lambda_2=0.25$ for subsequent experiments.

\vspace{-15pt}

\begin{figure}[h]
  \centering
  \includegraphics[width=0.95\linewidth]{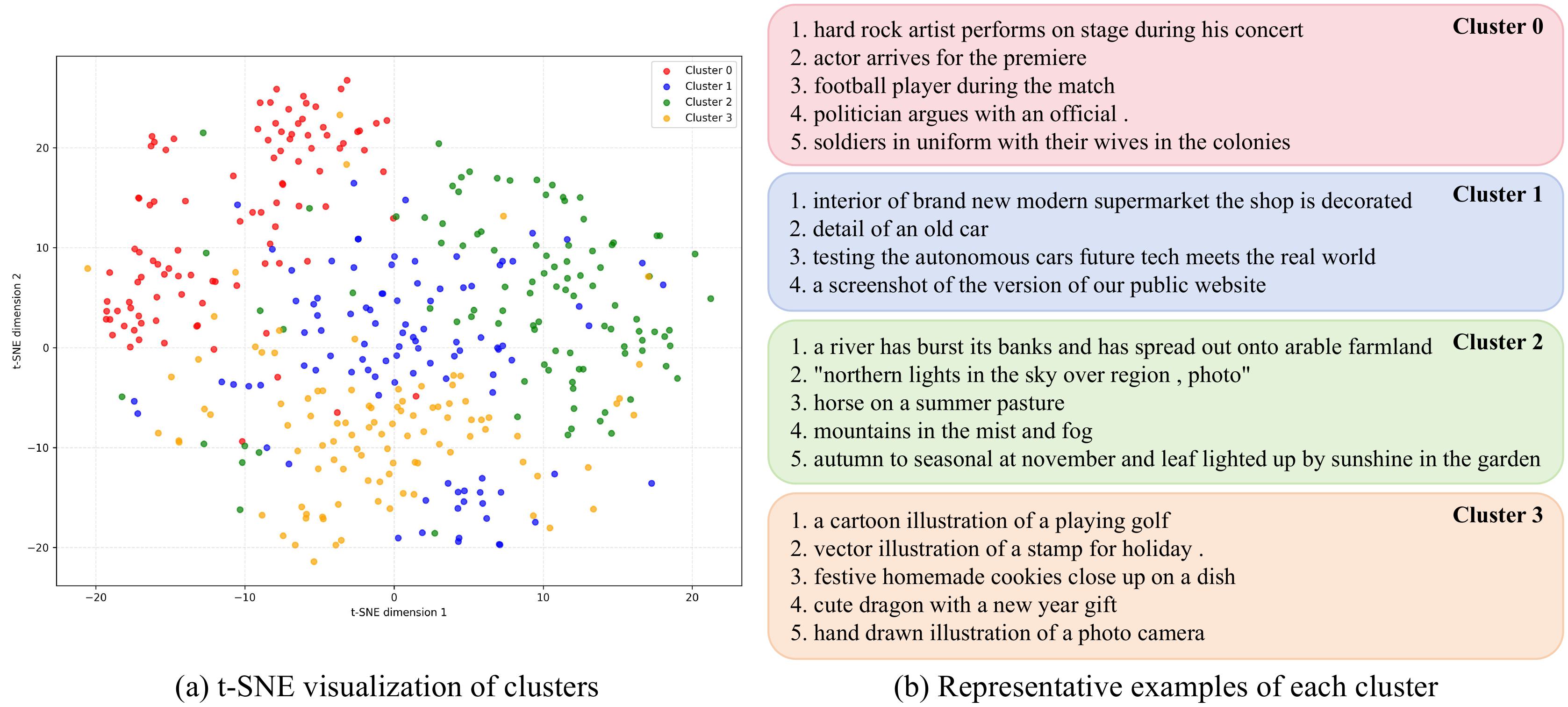}
  \caption{(a) t-SNE visualization of text embedding clusters and (b) representative caption examples of each semantic cluster}
  \label{fig:cluster}
  \vspace{-30pt}
\end{figure}

\subsection{Qualitative Analysis of Aspect Clusters}
\vspace{-5pt}
We analyze the clustering results in Fig.~\ref{fig:cluster} to verify aspect partitioning. Fig.~\ref{fig:cluster}(a) shows t-SNE text embedding clusters, and Fig.~\ref{fig:cluster}(b) lists representative samples of four semantic groups. The four clusters correspond to distinct semantic aspects: Cluster 0 covers human activities and events; Cluster 1 covers miscellaneous static objects and abstract concepts; Cluster 2 covers natural scenery; Cluster 3 covers artistic illustrations.

In the embedding space, Cluster 0, 2 and 3 form clearly separated regions due to their unified themes. By contrast, Cluster 1 involves diverse abstract semantics, resulting in a central distribution. These observations verify that our text-based clustering successfully groups samples with consistent descriptive aspects, laying a solid foundation for the proposed aspect-guided regularization.

\vspace{-10pt}

\section{Conclusion}
\vspace{-10pt}
In this paper, we introduce aspect-aware semantic clustering and propose an aspect-guided consistency regularization framework to optimize the CLIP representation space by respecting the one-to-many nature of visual-textual descriptions. Extensive experiments across downstream tasks demonstrate that AspectCLIP consistently outperforms the baselines,  higher consistency scores indicate that AspectCLIP learns a more structured and consistent representation space.

\noindent\textbf{Acknowledgement.} \ This work was supported by the Basic and Applied
Basic Research Foundation of Guangdong Province
(2024A1515012287), National Key R\&D Program of China (2023YFA1011601).
\vspace{-15pt}
%
%
%
%
\bibliographystyle{unsrt}
\bibliography{prcv2026references}





\end{document}